\documentclass{article}

     \PassOptionsToPackage{numbers, compress}{natbib}

     \usepackage[preprint]{want_neurips_2023}

\usepackage{multirow}
\usepackage[utf8]{inputenc} 
\usepackage[T1]{fontenc}    
\usepackage{hyperref}       
\usepackage{url}            
\usepackage{booktabs}       
\usepackage{amsfonts}       
\usepackage{nicefrac}       
\usepackage{microtype}      
\usepackage{xcolor}         
\usepackage{graphicx}       
\usepackage{caption}        
\usepackage{subcaption}     
\usepackage{amsmath}        
\usepackage{csvsimple}      
\usepackage{easyReview}     
\usepackage{float}          

\title{Deep Double Descent for Time Series Forecasting: Avoiding Undertrained Models}

\author{%
  Valentino Assandri \\
  Bosch Research North America\\
  Bosch Center for AI\\
  384 Santa Trinita Ave\\
  Sunnyvale, CA 94085, USA \\
  \And
  Sam Heshmati \\
  Bosch Research North America\\
  Bosch Center for AI\\
  384 Santa Trinita Ave\\
  Sunnyvale, CA 94085, USA \\
  \texttt{sam.heshmati@us.bosch.com} \\
  \AND
  Burhaneddin Yaman\\
  Bosch Research North America\\
  Bosch Center for AI\\
  384 Santa Trinita Ave\\
  Sunnyvale, CA 94085, USA \\
  \texttt{burhaneddin.yaman@us.bosch.com@us.bosch.com}  
    \And
    Anton Iakovlev \\
    Bosch Global Business Services \\
    Bregenzer Str. 26A \\
    70469 Stuttgart-Feuerbach, Germany \\
   \And
   Ariel Emiliano Repetur \\
   Bosch Research Germany\\
   Bosch Center for AI\\
   Robert-Bosch-Campus 1  \\
   71272 Renningen, Germany \\
   \texttt{arielemiliano.repetur@bosch.com} \\
}

\begin{document}

\maketitle

\begin{abstract}
Deep learning models, particularly Transformers, have achieved impressive results in various domains, including time series forecasting. While existing time series literature primarily focuses on model architecture modifications and data augmentation techniques, this paper explores the training schema of deep learning models for time series; how models are trained regardless of their architecture. We perform extensive experiments to investigate the occurrence of deep double descent in several Transformer models trained on public time series data sets. We demonstrate epoch-wise deep double descent and that overfitting can be reverted using more epochs. Leveraging these findings, we achieve state-of-the-art results for long sequence time series forecasting in nearly 70\% of the 72 benchmarks tested. This suggests that many models in the literature may possess untapped potential. Additionally, we introduce a taxonomy for classifying training schema modifications, covering data augmentation, model inputs, model targets, time series per model, and computational budget.
\end{abstract}

\section{Introduction}
\label{introduction}
Time series forecasting, particularly long sequence time series forecastig (LSTF), plays a crucial role in various domains, including finance \cite{no_double_descent_ts}, healthcare \cite{healthcare_forecasting}, and climate science \cite{healthcare_forecasting}, enabling data-driven decision-making and predictive modeling. Deep learning has been applied in a multitude of methods to address the challenges of time series forecasting \cite{DeepLearnTS_Survey, TrafoTS_Survey}. These deep learning models, particularly Transformers, have achieved remarkable performance in numerous other domains, including natural language processing (NLP) \cite{llama, BERT} and computer vision (CV) \cite{VisionTrafo}. While existing literature on time series forecasting primarily focus on architectural modifications \citep{TrafoTS_Survey, DeepLearnTS_Survey, EmpiricalSturyTSForecasting} and data augmentations \cite{DataAugTS_Survey}, recent advancements in the NLP space have demonstrated the significance of innovating the training schema \citep{kaplan_scaling_laws, deepmind_scaling_laws, instructGPT}. By drawing inspiration from these studies in NLP, we aim to start bridging the gap in the time series literature by improving \textit{how} models are trained, regardless of their architecture. The focus of the training schema innovations in this paper is the epoch-wise deep double descent phenomenon \cite{double_descent_transformers, when_deep_double_descent}.

The key contributions are summarised as follows:
\begin{itemize}
    \item We show that an epoch-wise deep double descent can occur on time series data from publicly available benchmarks, thus challenging the practice of early-stopping with low patience and few training epochs for LSTF and highlighting the need to be aware of this phenomenon.
    \item We performed extensive experiments on nine public benchmark datasets. 
    Our emperical studies show that many of the current Transformers for LSTF have more potential than what is currently believed by achieving new state-of-the-art performance for Transformers for LSTF: for example, we improve the MSE test loss of FEDformer-f on the ILI dataset regardless of the prediction length by at least 0.3 points and with an average relative improvement of 11\%.
    \item We additionally introduce a framework to categorize the variety of modeling techniques that can be found in deep learning for time series, including a more detailed overview of training schema modifications.
\end{itemize}

\section{Preliminaries}
\label{sec:preli}
Time series forecasting involves predicting future values of a time series $X\in D$ based on its past values, where $X$ is in a specific domain, normally with $D=\mathbb{R}$. More precisely, given a history of $p$ time steps from the time step $t$, we want to predict the value of $X$ for the next $h$ time steps; i.e., the input $X^i = (X_{t-p+1}, X_{t-p+2},	\ldots, X_{t-1} , X_{t} ) \in D^{p}$  is mapped to the output $X^o = (X_{t+1}, X_{t+2},	\ldots , X_{t+h-1} , X_{t+h} ) \in D^{h}$ by some function $f$ \cite{DeepLearnTS_Survey}.
In the case that the history size $p$ or the horizon size $h$ are very large, we refer to the problem as a \textit{long sequence} time series forecasting (LSTF) problem \cite{Autoformer}.

The LSTF problem is normally divided into two broad categories: univariate or multivariate. In the univariate case the output domain is 1 dimensional, normally $D=\mathbb{R}$, whereas in the multivariate case the domain $D$ is $n$ dimensional, normally $D=\mathbb{R}^n$, with $n \geq 2$ \cite{DeepLearnTS_Survey}.

\section{Background}
\subsection{Related Work}
\label{sec:related}

\textbf{Transition in Deep Learning Architectures for Time Series.} Long sequence time series forecasting has been impacted by many of the general developments in deep learning; models have generally speaking evolved from Feedforward Neural Networks (FNN) to Recurrent Neural Networks (RNN), then Long Short-Term Memory (LSTM) networks, and finally to Transformer architectures \cite{DeepLearnTS_Survey}. FNNs struggle with temporal dependencies, leading to RNNs with feedback loops for capturing long-range dependencies. Conventional RNNs face vanishing gradients, prompting LSTMs with memory cells and gating mechanisms \cite{DeepLearnTS_Survey}. LSTMs' sequential nature limits scalability, leading to parallel-processing Transformers with self-attention mechanisms \citep{Autoformer, Influenza, TrafoTS_Survey, ogTransformers}. Despite deep neural networks and Transformers' popularity, alternative approaches and classical methods like ARIMA remain effective for time series forecasting without the added complexity of deep learning \citep{DeepLearnTS_Survey, NLinear, deep_learning_no_good_ts, hyndman}.

\textbf{Transformers for Time Series Forecasting.}
In recent times, the application of Transformer-based models for time series analysis has gained traction \cite{TrafoTS_Survey}.
Several Transformer models have been designed by changing the Vanilla Transformer \citep{ogTransformers, Influenza} to tackle the long sequence time series forecasting problem, such as Informer \cite{Informer}, Non Stationary Transfomer \cite{NonStationaryTransformer}, Temporal Fusion Transformer \cite{TemporalFusionTransformer}, Autoformer \cite{Autoformer}, Pyraformer \cite{Pyraformer} and FEDformer \cite{FEDformer}. Many of these Transformer architecture modifications have been categorized in \cite{TrafoTS_Survey}. 

\textbf{Transformers Training Schemata for Time Series.} Whereas the aforementioned papers on time series all focus on \textit{model architecture} modifications, \cite{DataAugTS_Survey} focuses on data augmentation strategies for time series.
Some recent studies \cite{aTSisWorth64Words, TemporalFusionTransformer, timae} make use of training schema modifications as an addition to the architecture modifications they propose, but training schemata are not the focus of their work. 
In \cite{no_double_descent_ts} the authors focus on studying how scaling the number of parameters in a model affects performance when forecasting financial data using different model architectures. 

\textbf{Transformers Training Schemata in Other Domains.} In the realm of NLP and computer vision, on the other hand, various strategies have been employed to optimize the training schema of deep learning models. InstructGPT, for instance, utilizes reinforcement learning from human feedback (RLHF) to fine-tune existing GPT models, allowing it to follow user instructions more accurately \cite{instructGPT}. Another common approach involves pretraining models on vast amounts of unstructured data, which enables them to learn general features and representations that can later be fine-tuned for specific tasks \cite{BERT, llama}. In autonomous driving, \cite{scene_trafo} combines model architecture innovations with a masking training schema on the model targets to improve the performance of AI based planning. Additionally, recent research on scaling laws demonstrates that model performance can be significantly improved by increasing the size of models, data, and computational resources \cite{kaplan_scaling_laws, deepmind_scaling_laws, llama}. These strategies, combined or separately, have led to groundbreaking advancements in both NLP and computer vision, pushing the boundaries of what ML models can achieve.

Although there are extensive surveys and papers available on network modification \cite{TrafoTS_Survey, DeepLearnTS_Survey} and data augmentation \cite{DataAugTS_Survey} for time series transformers, the time series community would benefit from more research on training schemata for deep learning models for time series. 
We contribute to closing this gap by focusing on the deep double descent phenomenon in time series and specifically studying epoch-wise deep double descent.

\subsection{Epoch-Wise Deep Double Descent for Time Series}

\begin{figure}[h!]
  \centering
  \includegraphics[width=10.5cm,clip]{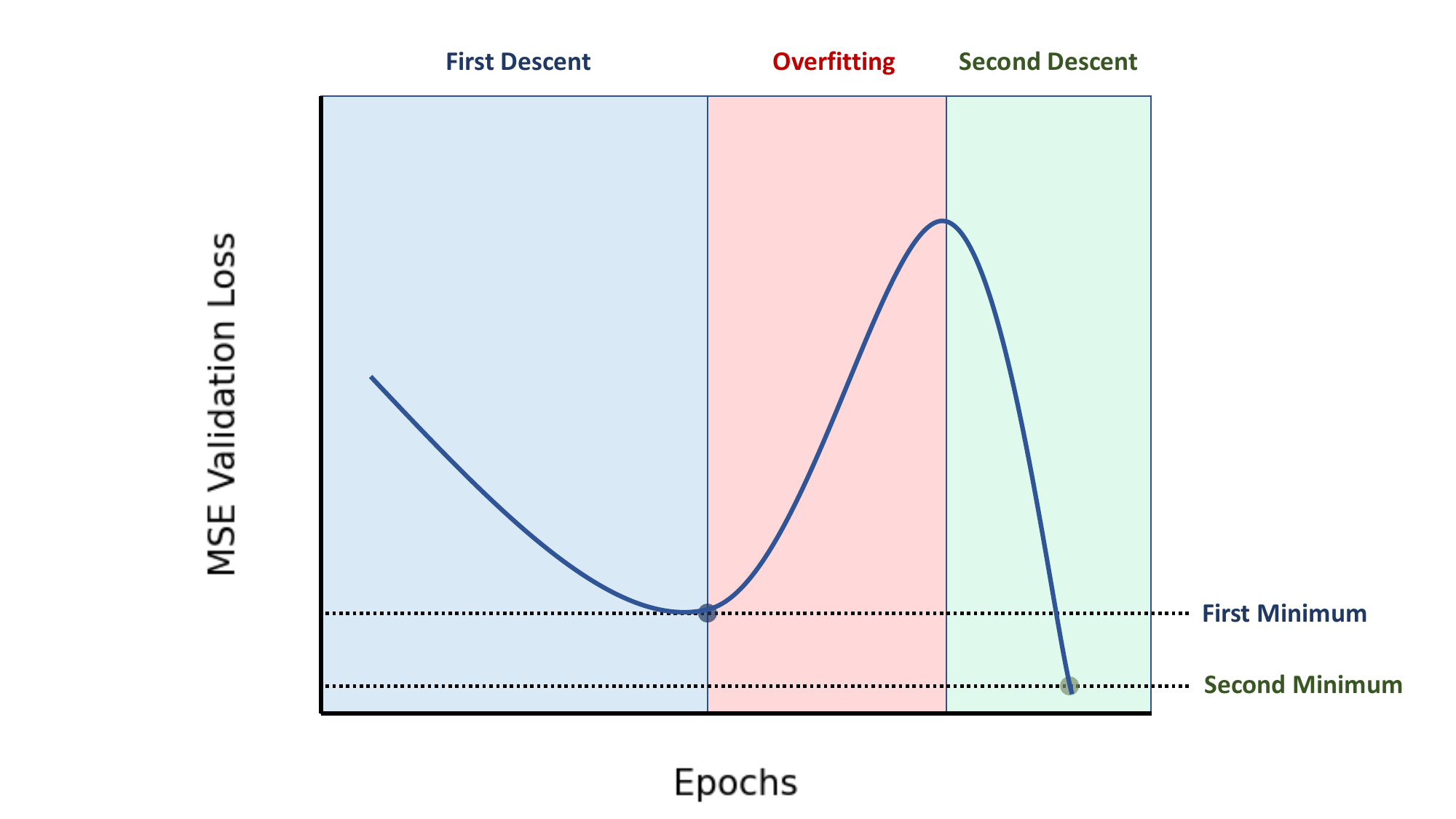}
  \caption{Figurative plot showing validation loss mapped against epochs. The validation initially drops during the "first descent" (light blue shading) and then starts to overfit as more epochs are added (light red shading). This overfitting continues until the model enters a second descent phase (light green shading). During the second descent the validation loss can drop below the previously achieved best performance.}
  \label{fig:labelled_dd_theory}
\end{figure}

Deep double descent, a recently re-popularized phenomenon in deep learning, challenges the conventional understanding of the relationship between model complexity, training epochs, and generalization performance \citep{compute_iso, double_descent_transformers, reconciling_ml_regimes, when_deep_double_descent}. 
Traditionally, it was believed that as model complexity increases, the model becomes prone to overfitting, and as a result, generalization performance would degrade; we refer to this as the classical regime \cite{double_descent_transformers}. However, the deep double descent phenomenon reveals a more nuanced pattern, where the generalization error first decreases, then increases, and then potentially decreases again as the model size and/or training epochs grow \cite{compute_iso}. In Figure \ref{fig:labelled_dd_theory} we see a figurative depiction of how a validation loss plot exhibiting deep double descent is expected to look.

This peculiar behavior suggests that large models, when trained for a sufficient number of epochs, can achieve better generalization performance despite the risk of overfitting.  In epoch-wise deep double descent we see the following pattern: the test loss as a function of epochs first decreases as the model learns, then increases as the model overfits the training data, until after a certain point it starts decreasing again \cite{double_descent_transformers}. In model wise deep double descent instead we plot the best achieved test loss against the number of parameters in the model \cite{double_descent_transformers}. 

The underlying theoretical principles that explain the occurrence and conditions of deep double descent remain less comprehensively understood \cite{when_deep_double_descent, compute_iso}. In \cite{double_descent_transformers} the authors postulate and experimentally show that noise in label data has an effect on whether the double descent phenomenon occurs; they show how for a standard computer vision benchmark (CIFAR), adding different levels of noise to the data leads to different levels of double descent. 

Time series forecasting community may gain significant advantages from further exploring and deepening the understanding of deep double descent in time series domain.
With the increasing use of deep learning techniques for forecasting noisy real-world time series data, investigating the insights of deep double descent can prove highly valuable.
We hence run experiments on public models and public data in Section \ref{sec:experiments} to see whether the double descent can also be seen in models trained on time series data.

\section{Experimental Evaluation and Discussion}
\label{sec:experiments}

To test whether a deep double descent can occur with time series forecasting, we perform a series of experiments on nine popular real-world datasets from different domains such as energy, economics, and health. The nine datasets (Electricity, Exchange, Traffic, Weather, ILI, ETTh1, ETTh2, ETTm1, ETTm2) were used as benchmarks in several Transformer papers \cite{FEDformer, Autoformer, Informer, NLinear}. As FEDformer exhibits the best performance on many of these benchmarks relative to the other Transformers we use it as the main baseline \cite{FEDformer}. We also include Informer \cite{Informer} and Autoformer \cite{Autoformer} to assess whether the double descent phenomenon is independent of model architecture.

\subsection{Experiment Setup}
To allow for a direct comparison, we take the multivariate experiments from \cite{NLinear} and increase the epoch count from 10 in the original experiments to 1000 and also increasing the patience from 3 to 1000. 
We note how the use of a low patience and a low epoch count is common among FEDformer \cite{FEDformer}, Autoformer \cite{Autoformer}, and Informer \cite{Informer}.
We hence run experiments on FEDformer, Informer and Autoformer on the same benchmark datasets from their respective studies using their hyperparameters. The bechmarks for the original training schema performance is taken from \cite{NLinear}.
As in \cite{FEDformer, Autoformer} the history
is 96 time steps for each experiment and different horizons are tested (96, 192, 336, and 720 time steps) on each dataset. 
For the ILI dataset we use the horizons 24, 36, 48, and 60 as in \cite{FEDformer}. The models were trained on an NVIDIA A100-80GB.

To accommodate the higher epoch count we make one hyperparameter change to the original setups from the respective papers: the learning rate decay. In the original papers the learning rate decays exponentially for all epochs \cite{NLinear, FEDformer}. In our experiments, the learning rate decays with the same rate as the original papers until epoch 5, after which it stays constant instead of further decaying exponentially. We underline how the models in \cite{FEDformer, Autoformer, Informer} were trained for less than 10 epochs and we hence mainly make changes to epochs that were not performed in their respective papers. No other hyperparameter tuning was done to reflect the increased epoch count. To limit the computational time, the patience for FEDformer on the Weather, ETTm1 and ETTm2 datasets was set to 200 instead of 1000. We underscore how no other hyperparamters were changed in these experiments compared to the original papers \cite{FEDformer, Informer, Autoformer} as the focus of this paper is the \textit{training schema} of the models.

\subsection{Main Results}
\label{sec:main_results}

Table \ref{tab:trafoOnly} shows an overview of the resulting test MSE and MAE on the benchmarks for each data set and prediction length; the table's rows represent each benchmark (data set and prediction length) and the columns are the different models. We highlight in bold the best results for each benchmark (row). The results indicate that employing a higher number of epochs often greatly improves the performance of the previously published techniques. 
For example, we improve the MSE test loss of FEDformer-f on the ILI dataset regardless of the prediction length by at least $0.3$ points and with an average relative improvement of 11\% . 
We underline how with this simple training schema modification we achieve state-of-the-art results for Transformers on a majority of the benchmarks. 

\setlength\tabcolsep{3pt} %
\begin{table}[]
\tiny
    \centering
   \begin{tabular}{|ll||ll|ll|ll||ll|ll|ll|}
    \hline
        \textbf{} & \textbf{} & \multicolumn{6}{c||}{\textbf{Double Descent}} & \multicolumn{6}{c|}{\textbf{Original}} \\ \hline 
        \textbf{} & \textbf{} & \multicolumn{2}{c|}{\textbf{FEDformer}} & \multicolumn{2}{c|}{\textbf{Autoformer}} & \multicolumn{2}{c||}{\textbf{Informer}} & \multicolumn{2}{c|}{\textbf{FEDformer}} & \multicolumn{2}{c|}{\textbf{Autoformer}}  & \multicolumn{2}{c|}{\textbf{Informer}} \\ \hline
         ~ & ~ & MSE & MAE & MSE & MAE & MSE & MAE & MSE & MAE & MSE & MAE & MSE & MAE \\ \hline
        \multirow{4}{*}{\rotatebox[origin=c]{90}{ETTh1}} & 96 & \textbf{0.376} &\textbf{ 0.413} & 0.436 & 0.448 & 0.933 & 0.764 & \textbf{0.376} & 0.419 & 0.449 & 0.459 & 0.865 & 0.713 \\ 
        ~ & 192 & 0.427 & 0.449 & 0.444 & 0.451 & 1.034 & 0.798 & \textbf{0.42} & \textbf{0.448} & 0.5 & 0.482 & 1.008 & 0.792 \\ 
        ~ & 336 & \textbf{0.448} & \textbf{0.462} & 0.516 & 0.493 & 1.098 & 0.819 & 0.459 & 0.465 & 0.521 & 0.496 & 1.107 & 0.809 \\ 
        ~ & 720 & \textbf{0.479} & \textbf{0.495} & 0.5 & 0.501 & 1.13 & 0.83 & 0.506 & 0.507 & 0.514 & 0.512 & 1.181 & 0.865 \\ 
        \multirow{4}{*}{\rotatebox[origin=c]{90}{ETTh2}} & 96 & \textbf{0.34} & \textbf{0.384} & 0.419 & 0.433 & 2.992 & 1.365 & 0.346 & 0.388 & 0.358 & 0.397 & 3.755 & 1.525 \\ 
        ~ & 192 & 0.431 & 0.44 & 0.45 & 0.448 & 7.184 & 2.206 & \textbf{0.429} & \textbf{0.439} & 0.456 & 0.452 & 5.602 & 1.931 \\ 
        ~ & 336 & 0.503 & 0.495 & \textbf{0.477} & \textbf{0.48 }& 5.662 & 1.926 & 0.496 & 0.487 & 0.482 & 0.486 & 4.721 & 1.835 \\ 
        ~ & 720 & 0.478 & 0.484 & 0.482 & 0.489 & 4.492 & 1.803 & \textbf{0.463} & \textbf{0.474} & 0.515 & 0.511 & 3.647 & 1.625 \\ 
        \multirow{4}{*}{\rotatebox[origin=c]{90}{ETTm1}} & 96 & \textbf{0.364} & 0.413 & 0.389 & \textbf{0.412} & 0.624 & 0.559 & 0.379 & 0.419 & 0.505 & 0.475 & 0.672 & 0.571 \\ 
        ~ & 192 & \textbf{0.406} & \textbf{0.435} & 0.465 & 0.456 & 0.717 & 0.615 & 0.426 & 0.441 & 0.553 & 0.496 & 0.795 & 0.669 \\ 
        ~ & 336 &\textbf{0.443} & \textbf{0.457} & 0.605 & 0.512 & 0.835 & 0.673 & 0.445 & 0.459 & 0.621 & 0.537 & 1.212 & 0.871 \\ 
        ~ & 720 & \textbf{0.523} & 0.492 & 0.75 & 0.571 & 0.904 & 0.714 & 0.543 & \textbf{0.49} & 0.671 & 0.561 & 1.166 & 0.823 \\ 
        \multirow{4}{*}{\rotatebox[origin=c]{90}{ETTm2}} & 96 & \textbf{0.189} & \textbf{0.282} & 0.244 & 0.317 & 0.409 & 0.48 & 0.203 & 0.287 & 0.255 & 0.339 & 0.365 & 0.453 \\ 
        ~ & 192 & \textbf{0.255} & \textbf{0.323} & 0.275 & 0.333 & 0.84 & 0.721 & 0.269 & 0.328 & 0.281 & 0.34 & 0.533 & 0.563 \\ 
        ~ & 336 & 0.327 & \textbf{0.365} & 0.336 & 0.37 & 1.44 & 0.921 & \textbf{0.325} & 0.366 & 0.339 & 0.372 & 1.363 & 0.887 \\ 
        ~ & 720 & 0.439 & 0.428 & 0.445 & 0.434 & 3.95 & 1.473 & \textbf{0.421} & \textbf{0.415} & 0.433 & 0.432 & 3.379 & 1.338 \\ 
        \multirow{4}{*}{\rotatebox[origin=c]{90}{Electricity}} & 96 & \textbf{0.188} & \textbf{0.304} & 0.201 & 0.315 & 0.333 & 0.412 & 0.193 & 0.308 & 0.201 & 0.317 & 0.274 & 0.368 \\ 
        ~ & 192 & \textbf{0.194} & \textbf{0.309} & 0.226 & 0.334 & 0.319 & 0.404 & 0.201 & 0.315 & 0.222 & 0.334 & 0.296 & 0.386 \\ 
        ~ & 336 & \textbf{0.211} & \textbf{0.324} & 0.219 & 0.331 & 0.327 & 0.409 & 0.214 & 0.329 & 0.231 & 0.338 & 0.3 & 0.394 \\ 
        ~ & 720 & 0.271 & 0.37 & 0.256 & 0.361 & 0.337 & 0.415 & \textbf{0.246} & \textbf{0.355} & 0.254 & 0.361 & 0.373 & 0.439 \\
        \multirow{4}{*}{\rotatebox[origin=c]{90}{Exchange}} & 96 & \textbf{0.125} & \textbf{0.254} & 0.179 & 0.31 & 0.903 & 0.787 & 0.148 & 0.278 & 0.197 & 0.323 & 0.847 & 0.752 \\ 
        ~ & 192 & \textbf{0.24} & \textbf{0.353} & 0.441 & 0.472 & 1.087 & 0.83 & 0.271 & 0.38 & 0.3 & 0.369 & 1.204 & 0.895 \\ 
        ~ & 336 & \textbf{0.402} & \textbf{0.465} & 1.08 & 0.778 & 1.44 & 0.986 & 0.46 & 0.5 & 0.509 & 0.524 & 1.672 & 1.036 \\ 
        ~ & 720 & 1.154 & 0.823 & \textbf{1.09} & \textbf{0.812} & 2.09 & 1.193 & 1.195 & 0.841 & 1.447 & 0.941 & 2.478 & 1.31 \\ 
        \multirow{4}{*}{\rotatebox[origin=c]{90}{Traffic}} & 96 & \textbf{0.578} & \textbf{0.358} & 0.63 & 0.402 & 0.718 & 0.399 & 0.587 & 0.366 & 0.613 & 0.388 & 0.719 & 0.391 \\ 
        ~ & 192 & 0.607 & 0.376 & 0.669 & 0.417 & 0.742 & 0.415 & \textbf{0.604} & \textbf{0.373} & 0.616 & 0.382 & 0.696 & 0.379 \\ 
        ~ & 336 & \textbf{0.619} & 0.38 & 0.703 & 0.436 & 0.775 & 0.429 & 0.621 & 0.383 & 0.622 & \textbf{0.337} & 0.777 & 0.42 \\ 
        ~ & 720 & 0.629 & \textbf{0.381} & 0.696 & 0.423 & 0.821 & 0.45 & \textbf{0.626} & 0.382 & 0.66 & 0.408 & 0.864 & 0.472 \\ 
        \multirow{4}{*}{\rotatebox[origin=c]{90}{Weather}} & 96 & 0.239 & 0.322 & 0.226 & 0.299 & 0.437 & 0.463 & \textbf{0.217} & \textbf{0.296} & 0.266 & 0.336 & 0.3 & 0.384 \\ 
        ~ & 192 & 0.284 & 0.342 & 0.317 & 0.381 & 0.508 & 0.499 & \textbf{0.276} & \textbf{0.336} & 0.307 & 0.367 & 0.598 & 0.544 \\ 
        ~ & 336 & 0.355 & 0.397 & 0.347 & 0.391 & 0.623 & 0.558 & \textbf{0.339} & \textbf{0.38} & 0.359 & 0.395 & 0.578 & 0.523 \\ 
        ~ & 720 & \textbf{0.392} & \textbf{0.406} & 0.449 & 0.45 & 0.982 & 0.75 & 0.403 & 0.428 & 0.419 & 0.428 & 1.059 & 0.741 \\ 
        \multirow{4}{*}{\rotatebox[origin=c]{90}{ILI}} & 24 & \textbf{2.628} & \textbf{1.06} & 3.563 & 1.259 & 5.246 & 1.585 & 3.228 & 1.26 & 3.483 & 1.287 & 5.764 & 1.677 \\ 
        ~ & 36 & \textbf{2.324} & \textbf{0.957} & 2.999 & 1.137 & 4.802 & 1.525 & 2.679 & 1.08 & 3.103 & 1.148 & 4.755 & 1.467 \\ 
        ~ & 48 & \textbf{2.31} & \textbf{0.98} & 3.082 & 1.163 & 5.041 & 1.562 & 2.622 & 1.078 & 2.669 & 1.085 & 4.763 & 1.469 \\ 
        ~ & 60 & \textbf{2.566} & \textbf{1.059} & 2.903 & 1.139 & 5.275 & 1.605 & 2.857 & 1.157 & 2.77 & 1.125 & 5.264 & 1.564 \\ \hline
    \end{tabular}

    \caption{Table of test loss from training FEDformer, Autoformer and Informer on the 9 standard benchmarks for LSTF comparing the original results (taken from \cite{FEDformer}) and comparing them to the results of the same models with the same hyperparameters but with more training epochs (labelled as "Double Descent"). We see how with a minor change to the training schema, i.e. increasing the epoch count and no longer performing early-stopping with low patience, we improve the test performance of Transformer-based LSTF models on a majority of the benchmarks. }
    \label{tab:trafoOnly}
\end{table}


\subsection{Discussion of Results}
As noted in Section \ref{sec:main_results}, the results taking advantage of a potential double descent normally outperform the previous implementations with only 10 epochs. We see how in around 62.5\% of the benchmarks, FEDformer's performance improves with the new training schema and there is on average a relative improvement of 6\% in the performance. In the cases where the performance does not improve, the performance only deteriorates by 3\%. Out of the 72 benchmarks (9 datasets with 4 prediction lengths each and 2 metrics) we set a new state-of-the-art 50 times or around on 69\% of benchmarks.

\begin{figure}
     \centering
     \begin{subfigure}[b]{0.328\textwidth}
         \centering
         \includegraphics[width=\textwidth]{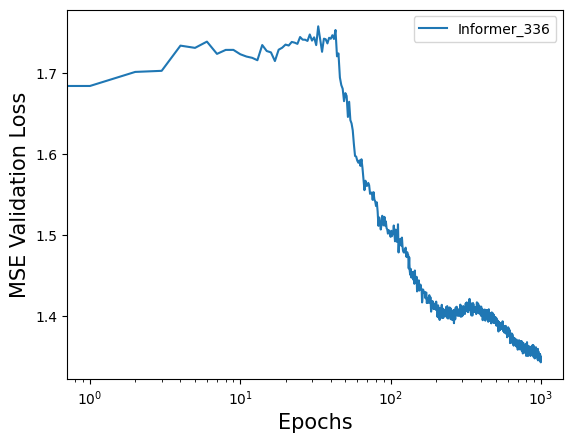}
         \caption{ETTh1}
         \label{fig:etth1Fig}
     \end{subfigure}
     \hfill
     \begin{subfigure}[b]{0.328\textwidth}
         \centering
         \includegraphics[width=\textwidth]{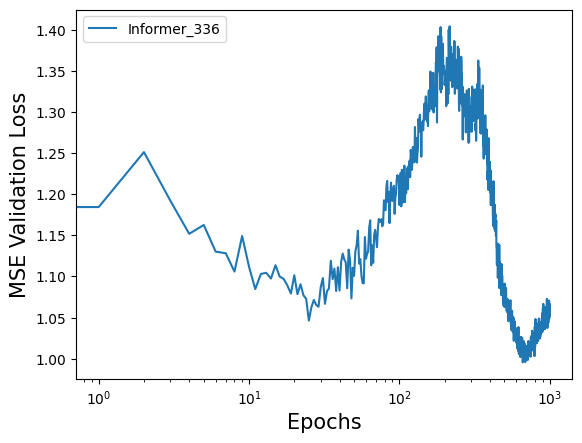}
         \caption{ETTh2}
         \label{fig:exchange} 
     \end{subfigure} 
     \begin{subfigure}[b]{0.328\textwidth}
         \centering
         \includegraphics[width=\textwidth]{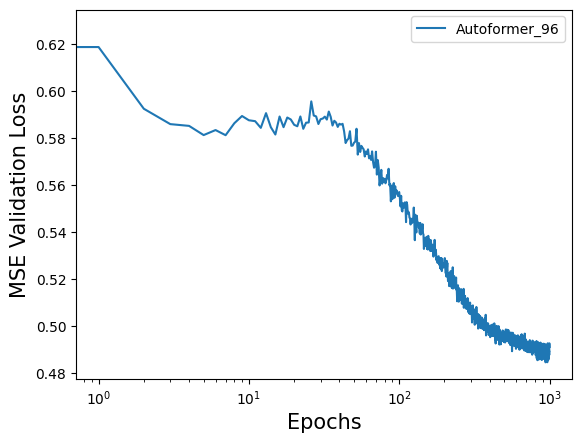}
         \caption{ETTm1}
         \label{fig:ettm1Fig}
     \end{subfigure}
     \hfill \\

     \begin{subfigure}[b]{0.328\textwidth}
         \centering
         \includegraphics[width=\textwidth]{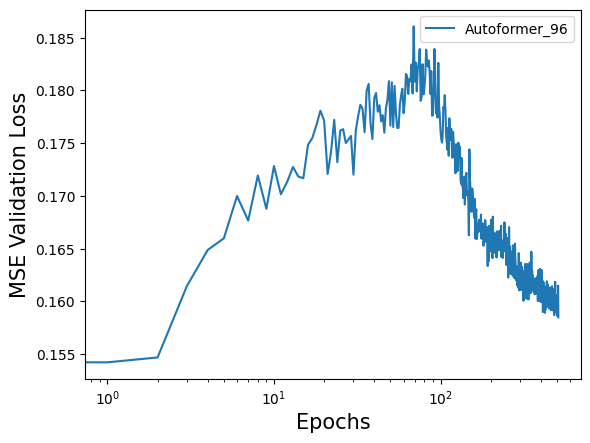}
         \caption{ETTm2}
         \label{fig:ettm2}
     \end{subfigure}
     \hfill
     \begin{subfigure}[b]{0.328\textwidth}
         \centering
         \includegraphics[width=\textwidth]{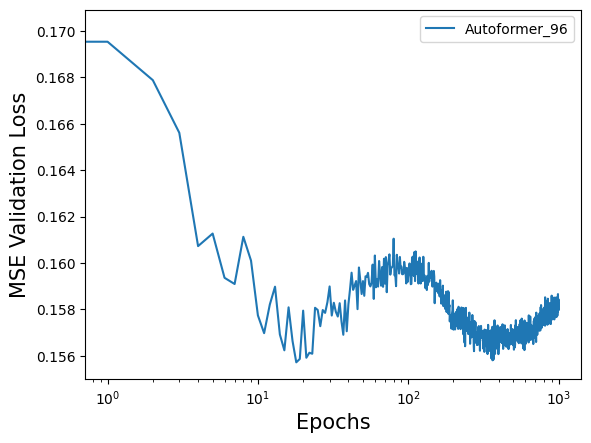}
         \caption{Electricity}
         \label{fig:ele} 
     \end{subfigure} 
     \begin{subfigure}[b]{0.328\textwidth}
         \centering
         \includegraphics[width=\textwidth]{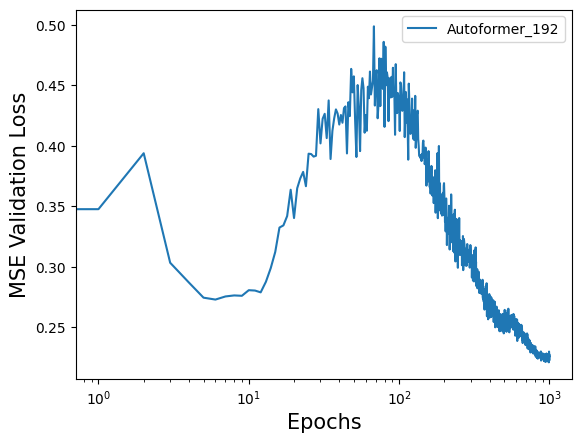}
         \caption{Exchange}
         \label{fig:exchange}
     \end{subfigure}
     \hfill \\

    \begin{subfigure}[b]{0.328\textwidth}
         \centering
         \includegraphics[width=\textwidth]{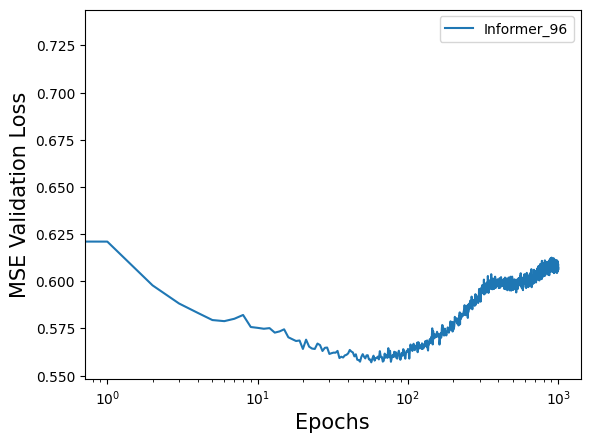}
         \caption{Traffic}
         \label{fig:traffic}
     \end{subfigure}
     \hfill
     \begin{subfigure}[b]{0.328\textwidth}
         \centering
         \includegraphics[width=\textwidth]{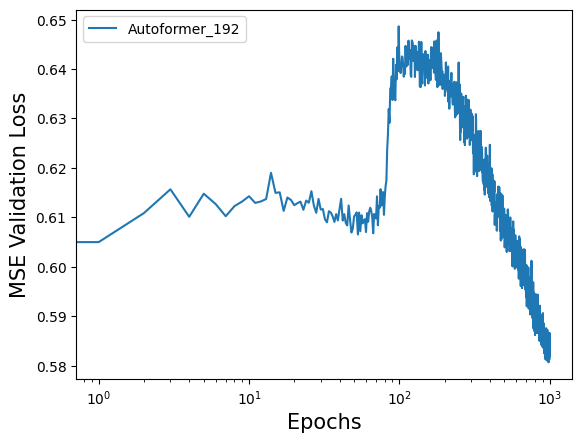}
         \caption{Weather}
         \label{fig:weatheer} 
     \end{subfigure} 
     \begin{subfigure}[b]{0.328\textwidth}
         \centering
         \includegraphics[width=\textwidth]{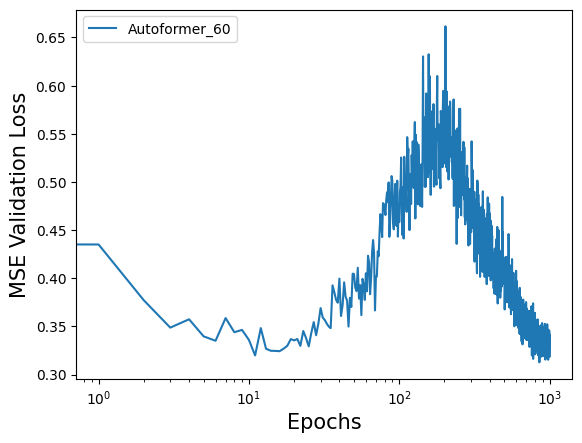}
         \caption{ILI}
         \label{fig:ili}
     \end{subfigure}
     \hfill 
     \caption{Validation loss of different Transformer based models showing examples of classical regime validation losses and modern regime double descentson different datasets. Note: the plots use a log scale in the epochs axis. We can see in the loss curves how the classical "U" shape expected is not always present when epochs are scaled enough and that overfitting is often a temporary phenomenon. }
    \label{fig:vali_loss}
\end{figure}

As  Figure \ref{fig:vali_loss} illustrates, many of the Transformer based models keep on improving (validation loss drops) in performance as the number of epochs is increased. We furthermore even see the emergence of epoch-wise double descent: for example, Autoformer on the Exchange dataset (Figure \ref{fig:exchange}) has a deteriorating validation loss from below 1 to over 2 up to around epoch 100 and it then drops below the previous low point achieved in the first "descent".

We see two reasons why we do not achieve increases in the performance on some experiments. Firstly, as we can see in Figure \ref{fig:traffic} with the Informer model on the Traffic data, models seem to be overfitting up to the 1000 epoch cut-off. 
We postulate that a double descent is still possible but occurs after too many epochs for it to be practical to train when considering the good performance achieved during the first descent. 
Secondly, we have double descents that remove the effects of overfitting but do not descend enough to find a new global minimum. 
This further highlights the need to view the double descent phenomenon as a factor to critically study during the training of a model.

\begin{figure}[H]
  \centering
  \includegraphics[width=12.8cm,clip]{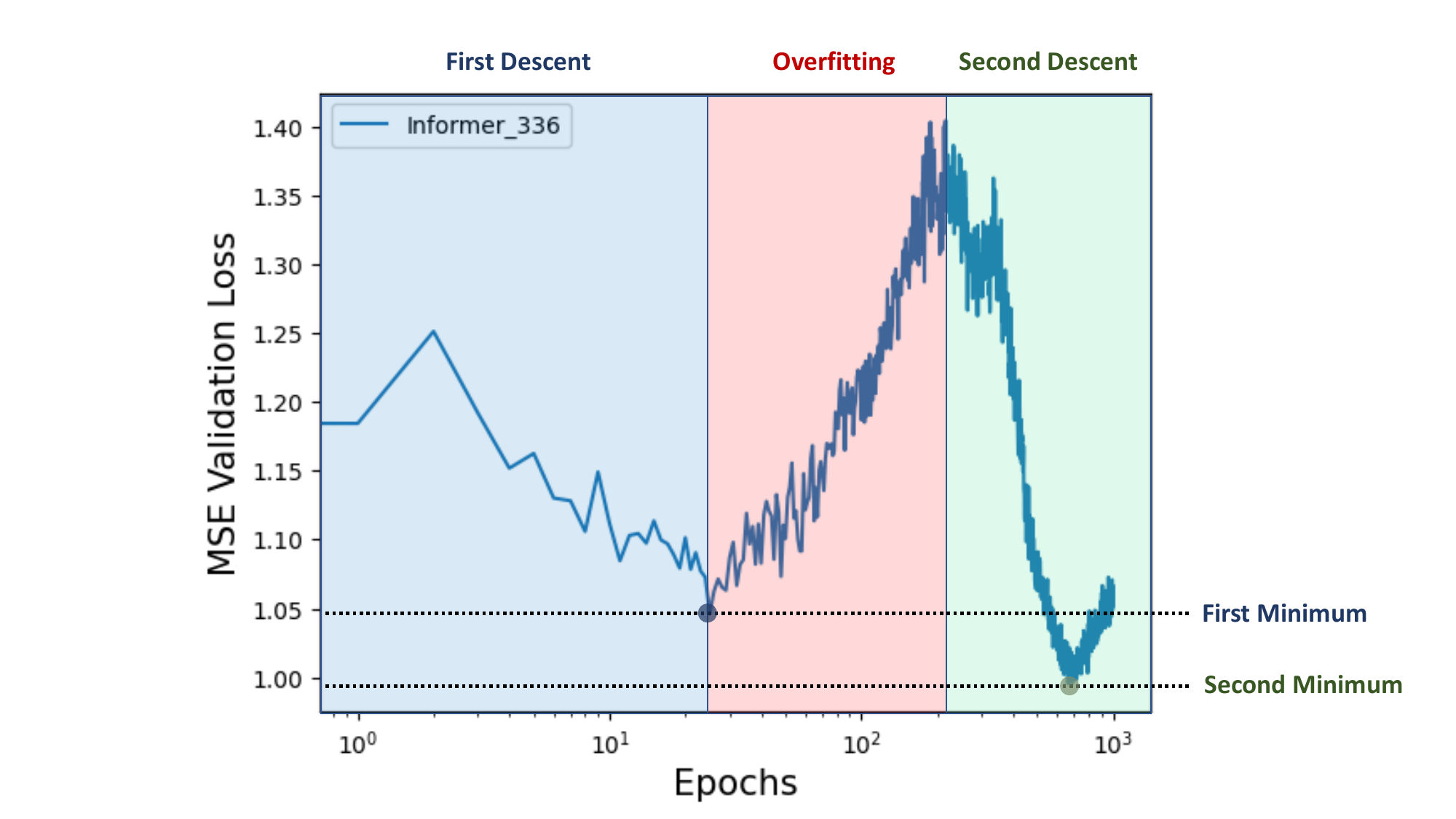}
  \caption{Validation loss of Informer with a prediction length of 336 time steps on the ILI dataset. We label in blue the initial descent, in red the epochs were overfitting occurs, and in green the second descent. We note the similarity of this plot from real data with the theoretical expectations shown in Figure \ref{fig:labelled_dd_theory}.}
  \label{fig:labelled_dd}
\end{figure}

In Figure \ref{fig:labelled_dd} we label the different phases of the loss curve for the ILI loss curve for Informer; differentiating between the first descent, the overfitting phase and the second descent.
Hence we stipulate that many of the Transformer models in \cite{FEDformer}, \cite{Autoformer} and \cite{Informer}, that normally use less than 10 epochs, have more potential than what was initially published.

\subsection{Limitations}
We acknowledge several aspects of our experiments and results that warrant further exploration or clarification. We selected the one thousand epoch cut-off out of computational constraints, which might not be the optimal choice for every application; for example, some applications might need only one hundred epochs while others ten thousand. More extensive testing and experimentation with different cut-offs could reveal more nuanced insights and improve the understanding of the epoch-wise deep double descent phenomenon in time series models. Additionally, it is important to note there are benchmarks where the state-of-the-art performance did not improve. While our findings highlight the potential of exploiting deep double descent, there may be cases where certain models or data sets do not exhibit the same benefits. Lastly, our study focused on three well-known models in the field (FEDformer \cite{FEDformer}, Autoformer \cite{Autoformer}, Informer \cite{Informer}), which does not constitute an exhaustive examination of all available time series models. We see our findings as a starting point for the field to explore the phenomenon in greater depth and to evaluate its potential across various applications and contexts.

\section{Taxonomy of Training Deep Learning Models for Time Series}
\label{sec:taxonomy}

\begin{figure}[h!]
  \centering
  \includegraphics[width=13.5cm,clip]{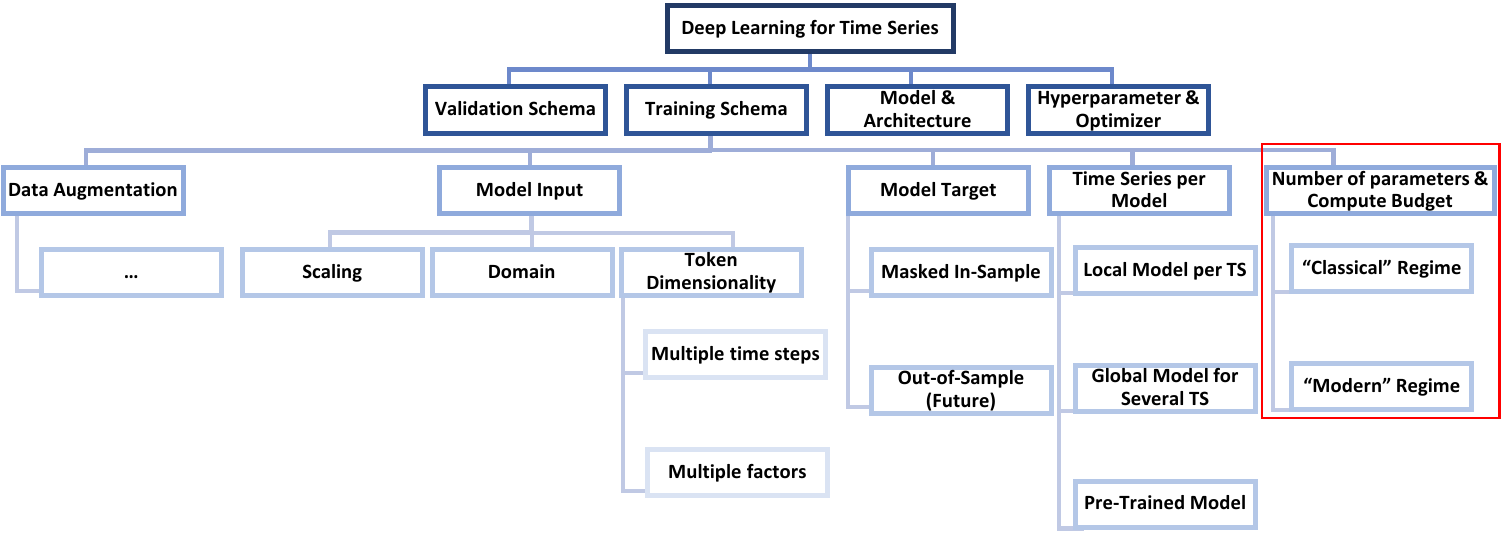}
  \caption{Taxonomy for Deep Learning for Time Series. We highlight which parts of the training schema were addressed in this paper.}
  \label{fig:taxonomy}
\end{figure}

In Section \ref{sec:experiments} we showed the promise of innovating on the training schemas of deep learning models for time series. As discussed in Section \ref{sec:related}, we consider that there is a gap in the current time series forecasting literature in regards to modifications other than architecture modifications. To start addressing this gap, we introduce a taxonomy specifically designed to categorize schemata in deep learning for time series, with a particular emphasis on training schema modifications. 
This classification system is visualized in Figure \ref{fig:taxonomy}. Our taxonomy aims to provide a clear framework for understanding and analyzing various approaches to enhance time series forecasting performance. 
At its highest level, the taxonomy differentiates between four types of modifications: \textit{i.} model and architecture modifications, \textit{ii.} hyperparameter and optimizer modifications, \textit{iii.} training schema modifications, and \textit{iv.} validation schema modifications. We limit the scope of this paper to the \textit{training schema modifications}. We organize modifications to training schemata differentiating between the following broad categories: model inputs, model targets,  time series per model, computational budget (number of parameters and compute) and data augmentation. We refer to \cite{DataAugTS_Survey} for a comprehensive survey of data augmentation techniques for time series.

\textbf{Model Target. }
Modifying the definition of the model targets is a training schema modification that can potentially improve the performance of deep learning models for time series forecasting. The main distinction that can be made here is between out-of-sample targets (i.e. target is the unseen future time steps \cite{DeepLearnTS_Survey}) and in-sample targets (i.e. the target is a group of time steps that can occur in the past \citep{EmpiricalSturyTSForecasting, timae}).
It is important to note that by model target modifications we do not mean transforming the \textit{target values of the time steps}, but rather choosing \textit{which time steps} are to be predicted. Out-of-sample target techniques include employing a rolling window and a non-rolling window for subsequent blocks \cite{EmpiricalSturyTSForecasting}. In-sample target techniques, on the other hand, can include strategies such as randomly hiding tokens or patches within the input sequence \citep{timae, EmpiricalSturyTSForecasting}. 

\textbf{Time Series per Model. }
Local models focus on learning and forecasting from individual time series, capturing patterns and trends specific to each series \cite{DeepLearnTS_Survey}. 
In contrast, global models are designed to learn from multiple time series concurrently, identifying common patterns and dependencies across series to improve generalization \cite{DeepLearnTS_Survey}.
By leveraging global models and incorporating techniques such as pre-training or zero-shot learning, researchers can harness shared knowledge across time series to potentially enhance forecasting performance and adapt more effectively to new or unseen time series data \citep{pretraining_for_TS_time_freq, zero_shot_TS_RNN}.
On the other hand, local models are tailored to the unique properties of the time series we wish to predict \cite{DeepLearnTS_Survey}.

We note the benchmarks defined in Informer \cite{Informer}, FEDformer \cite{FEDformer} and Autoformer \cite{Autoformer} are all benchmarks for local models. A famous example of global models and global model benchmarks is the research tackling the M4 and M5 competitions \cite{M4, M5}.

\textbf{Model Input. }
Several strategies have been proposed in the literature to augment model inputs. These strategies include altering input token dimensionality, such as representing one token for a single time step or employing patching techniques (i.e. one token represents multiple consecutive time steps \citep{aTSisWorth64Words, Dateformer}); selecting the input domain, for example in the time domain or frequency domain at the model level \cite{DataAugTS_Survey} or at the model component level \cite{FEDformer}; applying input scaling methods, such as normalizing the data or retaining the original scale \cite{NLinear}; and incorporating input extensions, which involve using raw input data as is and adding domain-specific knowledge, like time embeddings \citep{TemporalFusionTransformer, Signal2Vec}.

\textbf{Number of Parameters and Compute Budget. }
 Classical training regimes often prioritize simpler models and/or fewer training epochs, whereas modern regimes leverage larger models and extended training duration \cite{double_descent_transformers}.
 Additionally, examining the relationship between FLOPs (floating-point operations ) and model performance, rather than merely focusing on the number of training epochs, can provide valuable insights into the computational efficiency of the chosen training regime \cite{kaplan_scaling_laws, deepmind_scaling_laws}.

This taxonomy can be used to drive future research and identify other gaps in the time series forecasting literature for deep learning models.

\section{Conclusion}
\label{sec:conclusion}

\textbf{Contributions.} In conclusion, our study demonstrates that epoch-wise deep double descent can manifest in time series data from standard benchmarks, which calls into question the prevalent practice of early-stopping with low patience and limited training epochs when working with long sequence time series data. Furthermore, we attain new state-of-the-art performance for Transformers in long sequence time series forecasting and provide evidence that numerous existing state-of-the-art Transformer-based models for this task have more potential than what was currently believed. Lastly, we present a coherent framework to classify alterations in deep learning for time series and offer a more comprehensive overview of training schema modifications. This paper builds on, and is complementary to, the model and data augmentation literature in the field and its main goal is to highlight the potential of researching the training schema of deep learning models for time series.

\textbf{Broader Impacts} We briefly discuss the broader implications of our work, considering potential drawbacks. One concern is the environmental impact due to the increased computational requirements for training deep learning models for many more epochs as currently believed necessary, which could lead to a rise in CO2 emissions. Another consideration is that the growing reliance on computationally expensive training schemata may create barriers for researchers with limited access to computational resources or proprietary data, potentially leading to a disparity in research opportunities across the scientific community.

\textbf{Future Research Opportunities.} This study opens up new avenues for future research opportunities and encourages the exploration of innovative training schema techniques for time series forecasting tasks. We believe the following topics would be valuable contributions to the deep learning literature for time series field:  \textit{i.} whether the scaling laws seen in \cite{kaplan_scaling_laws} and \cite{deepmind_scaling_laws} for language models also hold in the time series domain, \textit{ii.} how pre-training time series models on large datasets affects global models, and \textit{iii.} statistically studying the properties of time series datasets and whether they can be linked to the shape of the validation loss curve.

\bibliographystyle{plain} 
\bibliography{refs} 


\end{document}